\documentclass[letterpaper]{article} 
\usepackage{colorai25}  
\usepackage{times}  
\usepackage{helvet}  
\usepackage{courier}  
\usepackage[hyphens]{url}  
\usepackage{graphicx} 
\usepackage{natbib}  
\usepackage{caption} 
\usepackage{algorithm}
\usepackage{algorithmic}

\usepackage{newfloat}
\usepackage{listings}
\DeclareCaptionStyle{ruled}{labelfont=normalfont,labelsep=colon,strut=off} 
\floatstyle{ruled}
\newfloat{listing}{tb}{lst}{}
\floatname{listing}{Listing}
\usepackage[]{todonotes}
\usepackage{amsfonts}
\usepackage{amsmath}
\usepackage{arydshln}
\usepackage{booktabs}
\usepackage{multirow}

\title{Low-Rank Adapters Meet Neural Architecture Search for LLM Compression
}
\author{
    J. Pablo Muñoz\textsuperscript{\rm 1}, Jinjie Yuan\textsuperscript{\rm 2}, Nilesh Jain\textsuperscript{\rm 1}
}
\affiliations{
    \textsuperscript{\rm 1}
    Intel Labs\\
    \textsuperscript{\rm 2}
    Intel Corporation \\

    \{pablo.munoz, jinjie.yuan, nilesh.jain\}@intel.com
}

\begin{document}

\maketitle

\begin{abstract}

The rapid expansion of Large Language Models (LLMs) has posed significant challenges regarding the computational resources required for fine-tuning and deployment. Recent advancements in low-rank adapters have demonstrated their efficacy in parameter-efficient fine-tuning (PEFT) of these models.
This retrospective paper comprehensively discusses innovative approaches that synergize low-rank representations with Neural Architecture Search (NAS) techniques, particularly weight-sharing super-networks. Robust solutions for compressing and fine-tuning large pre-trained models are developed by integrating these methodologies.  
Our analysis highlights the potential of these combined strategies to democratize the use of LLMs, making them more accessible for deployment in resource-constrained environments. The resulting models exhibit reduced memory footprints and faster inference times, paving the way for more practical and scalable applications of LLMs. Models and code are available at \url{https://github.com/IntelLabs/Hardware-Aware-Automated-Machine-Learning}.
\end{abstract}

%

\vspace{-0.3cm}
\section{Introduction and Preliminaries}

Structured low-rank representations \cite{KoBa09_tensordecomposition} have played a significant role in the latest successes in Artificial Inteligence (AI). For example, low-rank adaptation (LoRA) \cite{hu2022lora} is a preferred method for parameter-efficient fine-tuning (PEFT) of large foundation models \cite{Bommasani2021FoundationModels}. LoRA expands a linear layer by attaching low-rank adapters, $\boldsymbol{L_1}$ and $\boldsymbol{L_2}$, as demonstrated in the following equations:

\begin{equation}
    \boldsymbol Y = \boldsymbol X \boldsymbol W\label{eq:linear},
\end{equation}
\begin{equation}
    \boldsymbol Y = \boldsymbol X \boldsymbol W + s\boldsymbol X \boldsymbol L_1\boldsymbol L_2,
\label{eq:lora}
\end{equation}

where $\boldsymbol{X}$ is the input to the layer, and $\boldsymbol{W}$ are the layer's weights. \emph{s} is a scaling factor. $\boldsymbol{W}$ remains frozen, and only the adapters' weights are adapted during fine-tuning, which is significantly more efficient than performing full fine-tuning. Often, the number of parameters in the adapters is a minimal fraction of the total number of parameters in the base model. 

Neural Architecture Search (NAS) techniques attempt to identify a high-performing architectural configuration from a search space of candidate architectures \cite{nas1000}. NAS techniques evolved rapidly with the advent of \emph{deep learning}. However, many NAS techniques have become obsolete with the increasing size of large AI models because it is too resource-demanding to evaluate many possible architectures when models have billions of parameters. A particular efficient NAS technique relevant to this paper uses weight-sharing super-networks generated by activating substructures of the original neural network 
\cite{cai2020once}.  

We claim that the benefits of cross-pollination between low-rank representations and weight-sharing neural architecture search techniques are bi-directional: 

\begin{itemize}
    \item \textbf{NAS techniques enhance low-rank adapters, and,}
    \item \textbf{NAS becomes more efficient by incorporating the guidance of low-rank representations.}
\end{itemize}

In the following sections, we discuss several solutions that realize these benefits and suggest additional enhancements to be explored in the future.

\section{Elastic LoRA Adapters and Their Applications}

\begin{figure}[h]
\centering
\includegraphics[width=0.9\columnwidth]{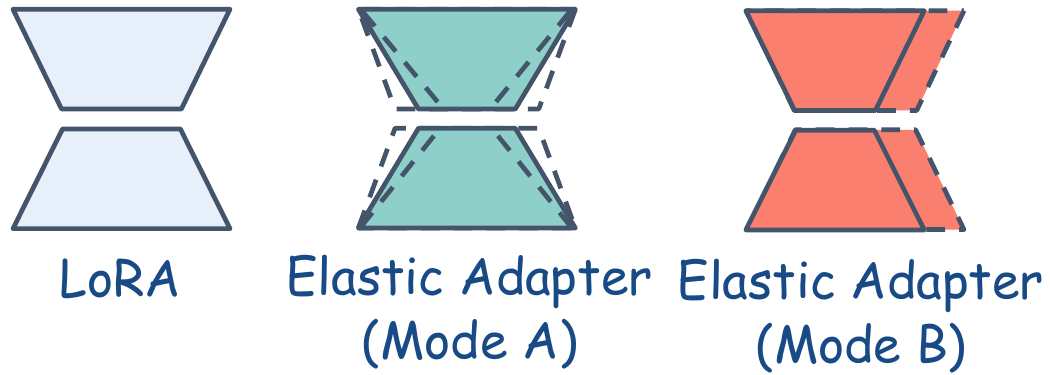}
\caption{Vanilla LoRA Adapter and two different modes of the elastic adapter. Mode A allows only the LoRA rank to be elastic, while Mode B also enables the input or output channels to be elastic.}
\label{fig:elastic-adapter}
\end{figure}

In this section, we first introduce the Elastic LoRA Adapter, highlighting its capability to dynamically adjust adapter configurations. This adaptability, coupled with an extensive sub-adapter search space, facilitates its application across various scenarios, enhancing model compression and fine-tuning efficiency and effectiveness. Next, let us delve step by step into the story of the combination of LoRA adapters and NAS techniques.

\paragraph{Elastic Adapter}

In weight-sharing NAS, an \emph{elastic} layer, as opposed to a traditional \emph{static} layer, has variable values for its properties. For instance, the weights, $\boldsymbol W \in \mathbb{R}^{m\times n}$, of a linear layer might be masked or sliced to activate a smaller structure, $\boldsymbol W \in \mathbb{R}^{m\times k}$ where $k < n$. By allowing the activation of variable configurations of a layer during the forward and backward passes, one is effectively training a super-network in which the smaller structures share their weights with their bigger counterparts.   
Recent advancements in NAS weight-sharing techniques have been utilized in conjunction with low-rank representations. 
As illustrated in Figure \ref{fig:elastic-adapter}, the Elastic LoRA Adapter primarily operates in two modes:

\textbf{i) Mode A}: In the LoRA adapter, matrices $\boldsymbol {L}_1 \in \mathbb{R}^{m \times r}$ and $\boldsymbol {L}_2 \in \mathbb{R}^{r \times n}$ can be rendered elastic by adopting smaller rank values. Specifically, $\boldsymbol L_{\delta 1} \in \mathbb{R}^{m \times \{r_0, r_1, ..., r\}}$ and $\boldsymbol {L}_{\delta 2} \in \mathbb{R}^{\{r_0, r_1, ..., r\} \times n}$, where $r_i \le r$ , $r \ll m$ and $r \ll n$ \cite{munoz-etal-2024-sqft}.

\textbf{ii) Mode B}: Alternatively, $\boldsymbol {L}_1$ and $\boldsymbol {L}_2$ can achieve elasticity by allowing the activation of substructures with reduced channel widths. This is represented as $\boldsymbol L_{\delta 1} \in \mathbb{R}^{\{m_0, m_1, ..., m\} \times \{r_0, r_1, ..., r\}}$ and $\boldsymbol {L}_{\delta 2} \in \mathbb{R}^{\{r_0, r_1, ..., r\} \times \{n_0, n_1, ..., n\}}$, where $m_i \le m$ and $n_i \le n$ \cite{munoz-etal-2024-lonas}. 

To this end, we will describe several methodologies in which low-rank structures and weight-sharing super-networks techniques can benefit each other. 

\subsection{Efficient Neural Architecture Search with the Guidance of Low-Rank Adapters}

\begin{figure}[H]
\centering
\includegraphics[width=0.9\columnwidth]{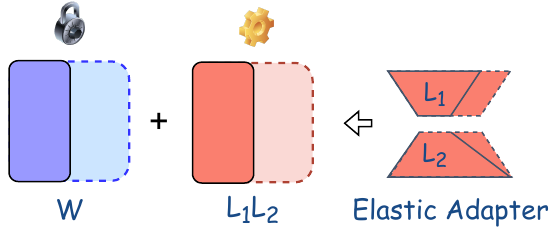} 
\caption{Elastic adapters guide the removal of elements in the frozen model weights, resulting in smaller, high-performing models. This process exemplifies the application of Mode B as depicted in Figure \ref{fig:elastic-adapter}.}
\label{fig:lonas}
\end{figure}

\paragraph{LoNAS} During fine-tuning, the sub-adapters activated can be used to guide the activation of substructures in the base model, as illustrated in Figure \ref{fig:lonas}. 
This approach corresponds to Mode B in Figure \ref{fig:elastic-adapter}, which is characterized by its ability to reduce the overall number of parameters in the model compared to Mode A. 
In this scenario, the adapters are elastic, and the frozen weights of the base model, denoted as $ \boldsymbol W \in \mathbb{R}^{m \times n}$ , are transformed into $\boldsymbol W_\delta \in \mathbb{R}^{m \times \{n_0, n_1, ..., n\}}$ or $ \boldsymbol W \in \mathbb{R}^{m \times n}$ into $\boldsymbol W_\delta \in \mathbb{R}^{\{m_0, m_1, ..., m\} \times n}$.The search space of possible low-rank adapter configurations is generated by allowing several configurations in the width and rank of the adapters.  
This alignment, as proposed by LoNAS \cite{munoz-etal-2024-lonas}, results in models with fewer parameters than the base model while maintaining a minimal drop in accuracy and achieving immediate improvements in inference speedup. 
LoNAS is analogous to traditional NAS, but with the key distinction that only the adapter parameters are trained. The expectation is that this training will guide and adapt the model's search and pruning processes. Due to the high cost of searching LLM with adapters, LoNAS also proposes heuristic sub-networks (i.e., middle point of the search space) to quickly evaluate the quality of the trained super-network. Users can then decide whether further search is necessary based on specific needs. This heuristic strategy has also been applied in subsequent works such as Shears \cite{munoz-etal-2024-shears} and SQFT \cite{munoz-etal-2024-sqft}, which are discussed later.

Empirical results demonstrate that this approach yields an inference speedup of up to 1.4x and can reduce the model parameters by approximately 80\% compared to the original model. 
For more detailed information, refer to Table \ref{tab:lonas_performance} and the subsequent section.

LoNAS enhancements have recently been proposed
\cite{sukthanker2024large} by applying elastic LoRA adapters to all the weight matrices of the transformer and allowing the removal of entire transformer blocks. Initially, LoNAS focused solely on the Self-Attention and MLP layers. While LoNAS and its extensions have proven effective, they still face challenges due to the size of state-of-the-art pre-trained models, which have driven the development of more efficient solutions, which we will discuss in the following sections.

\subsection{Restricting the Elasticity to the Adapter Rank and Exploiting Model Sparsity and Low Numerical Precision}

\begin{figure}[h]
\centering
\includegraphics[width=0.95\columnwidth]{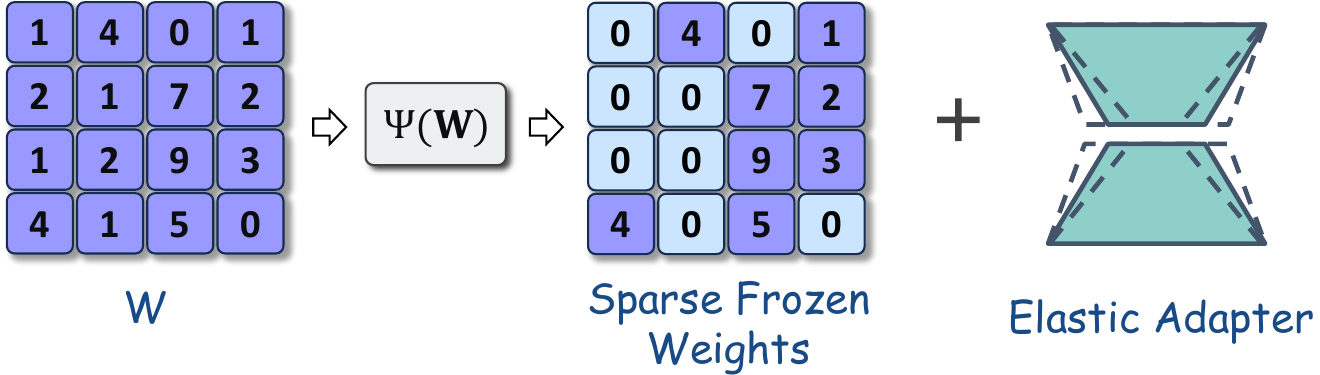} 
\caption{Elastic low-Rank adapters for fine-tuning sparse efficient models. This style exemplifies the application of Mode A as depicted in Figure \ref{fig:elastic-adapter}.}
\label{fig:shears}
\end{figure}

\paragraph{Shears} Building on LoNAS, Shears \cite{munoz-etal-2024-shears} proposes several modifications to enhance the efficiency of the fine-tuning stage. This approach constrains the application of elasticity exclusively to the low-rank structures, leaving the significantly demanding weights of the base model intact. This strategy is termed \textbf{Neural Low-Rank Adapter Search (NLS)}. Additionally, as illustrated in Figure \ref{fig:shears}, the base model can be sparsified using an arbitrary metric, $\Psi$, to determine the importance of the pretrained weights. A popular weight importance metric is Wanda \cite{sun2023wanda}, where a few feature input activations, $\boldsymbol X$, are used to assess the importance of the weights, i.e., $\Psi(\boldsymbol{W}) = |\boldsymbol W| \cdot \|\boldsymbol X\|_2$. This score, combined with a desired sparsity level, $s$, is used to obtain the sparse weights, $\boldsymbol{W}^p$, with a sparsity pattern $S\{\boldsymbol{W}^p\} = \{(i, j) \mid \boldsymbol{W}^p_{i, j} \not= 0, 1 \leq i \leq m, 1 \leq j \leq n\}$, such that $\lvert S\{\boldsymbol{W}^p\}\rvert \leq \lvert S\{\boldsymbol{W}\}\rvert$.

Overall, Shears introduces the concept of Mode A elastic adapters, which allow the rank values to be flexible, thereby enabling the exploration of more potential sub-adapters. This approach has been demonstrated to outperform traditional LoRA (with fixed rank values) and alleviates the challenge of setting the hyperparameter rank value when using LoRA. Most importantly, Shears found that the NLS algorithm is particularly well-suited for sparse models. When pre-trained weights are sparsified, there is a natural and significant drop in accuracy compared to dense models. On this basis, using NLS for fine-tuning can maximally recover or adapt the model's performance to a specific downstream task.

\paragraph{SQFT} Shears is extended by SQFT \cite{munoz-etal-2024-sqft} to manipulate sparse models on low numerical precision. SQFT is inspired by QLoRA \cite{dettmers2023qlora}, which was proposed to improve fine-tuning efficiency when using low-rank adapters. SQFT enables three different pipelines that account for the varying characteristics of the base models, such as whether they possess sparsity or have been quantized to low numerical precision. 

Empirical results demonstrate that by combining elastic LoRA adapters into the sparse or quantized base model, Shears, and SQFT enable effective fine-tuning of compressed models to adapt to specific downstream tasks. This approach produces compressed models that either improve or exhibit only minor drops in accuracy. The enhanced sparsity and precision can lead to significant speedups when utilizing runtimes optimized for these patterns.

However, a significant challenge in Shears and SQFT when dealing with compressed models and dense adapters is the potential limitations encountered when attempting to merge the low-rank adapters with the based model after fine-tuning. For instance, if the model is sparse but the adapters are dense, the sparsity in the model will be lost when merging. A similar limitation arises when the based model has a different numerical precision than the low-rank adapters used for fine-tuning. In the next section, we describe how these limitations are addressed to ensure the integrity and performance of the fine-tuned models. 

\subsection{Addressing the Challenges of Merging Adapters with Low-precision Sparse Models}

Within SQFT, two strategies, \emph{SparsePEFT} and \emph{QA-SparsePEFT}, are proposed to address the limitations described in the previous section. These limitations arise when attempting to merge the low-rank adapters with base models with differing sparsity patterns or numerical precision. The following sections provide a detailed discussion of these strategies. 

\paragraph{SparsePEFT} This strategy ensures that the sparsity in low-rank adapters is aligned with their corresponding base model's weights during fine-tuning. SQFT achieves this by generating a binary mask $\boldsymbol{M}$ for each weight matrix $\boldsymbol{W}$ in the base model. The mask $\boldsymbol{M}$ is $\forall i \forall j  (W_{i, j} \not= 0 \Rightarrow \boldsymbol{M}_{i, j} = 1)$. Utilizing $\boldsymbol{M}$, SQFT sparsifies the adapters' matrix ($\boldsymbol{L}_1\boldsymbol{L}_2$) to obtain $\boldsymbol{L}^p$, i.e., $\boldsymbol{L}^p = (\boldsymbol{L}_1\boldsymbol{L}_2) \odot \boldsymbol{M}$. This approach ensures sparsity awareness during fine-tuning, allowing the merging of the base model's weights and adapter's weights without losing the sparsity induced before fine-tuning.

\paragraph{QA-SparsePEFT} This strategy is employed by SQFT when the model has been quantized to lower numerical precision, and low-rank adapters with higher numerical precision are applied for fine-tuning. Quantization-aware SparsePEFT (QA-SparsePEFT) leverages the frozen zeros $\boldsymbol{z}$ and scales $\boldsymbol{s}$ resulting from the pre-fine-tuning stage in which each weight matrix, $\boldsymbol{W}$, was asymmetrically quantized. By utilizing $\boldsymbol{z}$, $\boldsymbol{s}$, the numerical target range $[0, 2^{n-1}]$ for quantization (where $n$ represents the target bit-width), and the pre-quantized sparse weights, $\boldsymbol{W}^p$, QA-SparsePEFT achieves quantization-aware fine-tuning with low-rank adapters on $\widehat{\boldsymbol{W}}^p_m$, i.e., the sparse quantized (merged) weights. This process is formalized as, 

\begin{equation}
\small
   \widehat{\boldsymbol{W}}^p_m = \text{clamp}\left( \text{round}\left( \frac{\boldsymbol{W^p+L^p}}{\boldsymbol{s}} \right) + \boldsymbol{z},0,2^{n}-1\right),
\end{equation}

To obtain the dequantized weights,$\tilde{\boldsymbol{W}}^p_m$, the inverse process is followed, 
\begin{equation}
\small
   \tilde{\boldsymbol{W}}^p_m = \boldsymbol{s} \left( \widehat{\boldsymbol{W}}^p_m - \boldsymbol{z} \right).
\end{equation}

In summary, the integration of Elastic LoRA Adapters with NAS techniques offers a promising approach to model compression and fine-tuning. By leveraging the flexibility of elastic adapters and the efficiency of NAS, methods like LoNAS, Shears, and SQFT demonstrate significant improvements in parameter reduction and inference speedup without sacrificing accuracy. Additionally, strategies such as SparsePEFT and QA-SparsePEFT address the challenges of merging (elastic or static) LoRA adapters with low-precision sparse models, ensuring robust performance and maintaining model integrity. These advancements highlight the potential of combining low-rank adapters with NAS to optimize large language models.

\begin{table*}[ht]
\setlength{\belowcaptionskip}{5pt}
\setlength{\tabcolsep}{4.5pt}
\caption{The performance of LoNAS using elastic adapters mode B, including accuracy score and model compression efficiency when fine-tuning LLaMA-7B on 15k unified commonsense reasoning dataset from LLM-Adapters \cite{hu2023llm_adapters}.
The average score represents the results across eight commonsense tasks. These results are reproduced from \citet{munoz-etal-2024-lonas}}
\small
\centering
\renewcommand\arraystretch{1.2}
\begin{tabular}{@{}lccccccccccc@{}}

\toprule
\textbf{Method} & \textbf{Average Score}  & \textbf{Relative Score} & \textbf{Total Params.} & \textbf{TFLOPs} & \textbf{Inference Speedup}  \\
\midrule
LoRA & 65.8 & 100.0\% & 6.7B & 1.7 & 1.00$\times$  \\
\cdashline{1-12}
\textbf{LoNAS (Heuristic Subnet)} & 65.2 & 99.1\% & 5.6B & 1.4 & 1.23$\times$  \\
\textbf{LoNAS (Search Subnet-1)} & 67.1 & 102.0\% & 5.6B & 1.4 & 1.28$\times$  \\
\textbf{LoNAS (Search Subnet-2)} & 65.6 & 99.7\% & 5.1B & 1.3 & 1.41$\times$  \\
\bottomrule

\end{tabular}
\label{tab:lonas_performance}
\end{table*}

\begin{table*}[htb]
\setlength{\belowcaptionskip}{5pt}
\setlength{\tabcolsep}{4.5pt}
\caption{The performance of Shears and SQFT from \citet{munoz-etal-2024-lonas}, when fine-tuning Mistral-7B-v0.3 on GSM8K using elastic adapters mode A.}
\small
\centering
\renewcommand\arraystretch{1.2}
\begin{tabular}{lcccc}

\toprule
\multirow{2}{*}{\textbf{Method}} & \multirow{2}{*}{\textbf{Accuracy}}  & \multirow{2}{*}{\textbf{Relative Acc.}} & \multirow{2}{*}{\textbf{Sparsity}} & \textbf{Precision} \\
&  & & & \textbf{(Base + Adapter / Base)} \\
\midrule
w/o tune & 36.0 & - & - & FP16  \\
LoRA & 44.1 & 100.0\% & 50\% & FP16  \\
\cdashline{1-5}
\textbf{Shears} & 45.1 & 102.3\% & 50\% & FP16 + FP16  \\
\textbf{SQFT + SparsePEFT} & 50.1 & 113.6\% & 50\% & FP16  \\
\textbf{SQFT} & 44.5 & 100.9\% & 50\% & INT4 + FP16  \\
\textbf{SQFT + QA-SparsePEFT} & 44.0 & 99.8\% & 50\% & INT4  \\
\bottomrule

\end{tabular}
\label{tab:sqft_performance}
\end{table*}  

\section{Performance Summary and Additional Considerations} 
 
We summarize the performance, accuracy, and compression efficiency of LoNAS (Table \ref{tab:lonas_performance}), SQFT (Table \ref{tab:sqft_performance}) and Shears (Tables \ref{tab:sqft_performance}, and \ref{tab:shears_performance}) from their respective papers. The reader can find additional details and an exhaustive list of experiments in each solutions' source. From these tables, we can observe that LoNAS can obtain competitive results compared to vanilla LoRA. However, LoNAS application is costlier than the other two discussed solutions due to the elasticity enabled in the model's weights, in addition to the inserted elastic adapters, which makes the fine-tuning stage more expensive. Shears and SQFT, on the other hand, are more fine-tuning efficient since their manipulation is only at the adapters' level.

\begin{table}[htb]
\setlength{\belowcaptionskip}{5pt}
\setlength{\tabcolsep}{4.5pt}
\caption{The performance of Shears for LLaMA-13B on a 10k unified math reasoning dataset from LLM-Adapters \cite{hu2023llm_adapters} using elastic adapters mode A.
These results are reproduced from \citet{munoz-etal-2024-lonas}, and the average score represents the results across four math tasks (GSM8K \cite{cobbe2021training_GSM8K}, AQUA \cite{ling-etal-2017-program-aqua}, MAWPS \cite{lan2022mwptoolkit} and SVAMP \cite{patel-etal-2021-nlp-svamp}).
}
\small
\centering
\renewcommand\arraystretch{1.2}
\begin{tabular}{lcccc}

\toprule
\multirow{2}{*}{\textbf{Method}} & \textbf{Average}  & \textbf{Relative} & \multirow{2}{*}{\textbf{Sparsity}} & \textbf{Non-zero}  \\
 & \textbf{Score}  & \textbf{Score} & & \textbf{Params.}  \\
\midrule
LoRA & 51.1 & 100.0\% & - & 13.0B  \\
\cdashline{1-5}
\textbf{Shears} & 52.0 & 101.8\% & 40\% & 8.0B  \\
\textbf{Shears} & 50.9 & 99.6\% & 50\% & 6.7B  \\
\bottomrule

\end{tabular}
\label{tab:shears_performance}
\end{table}

\begin{figure}[H]
\centering
\includegraphics[width=0.80\columnwidth]{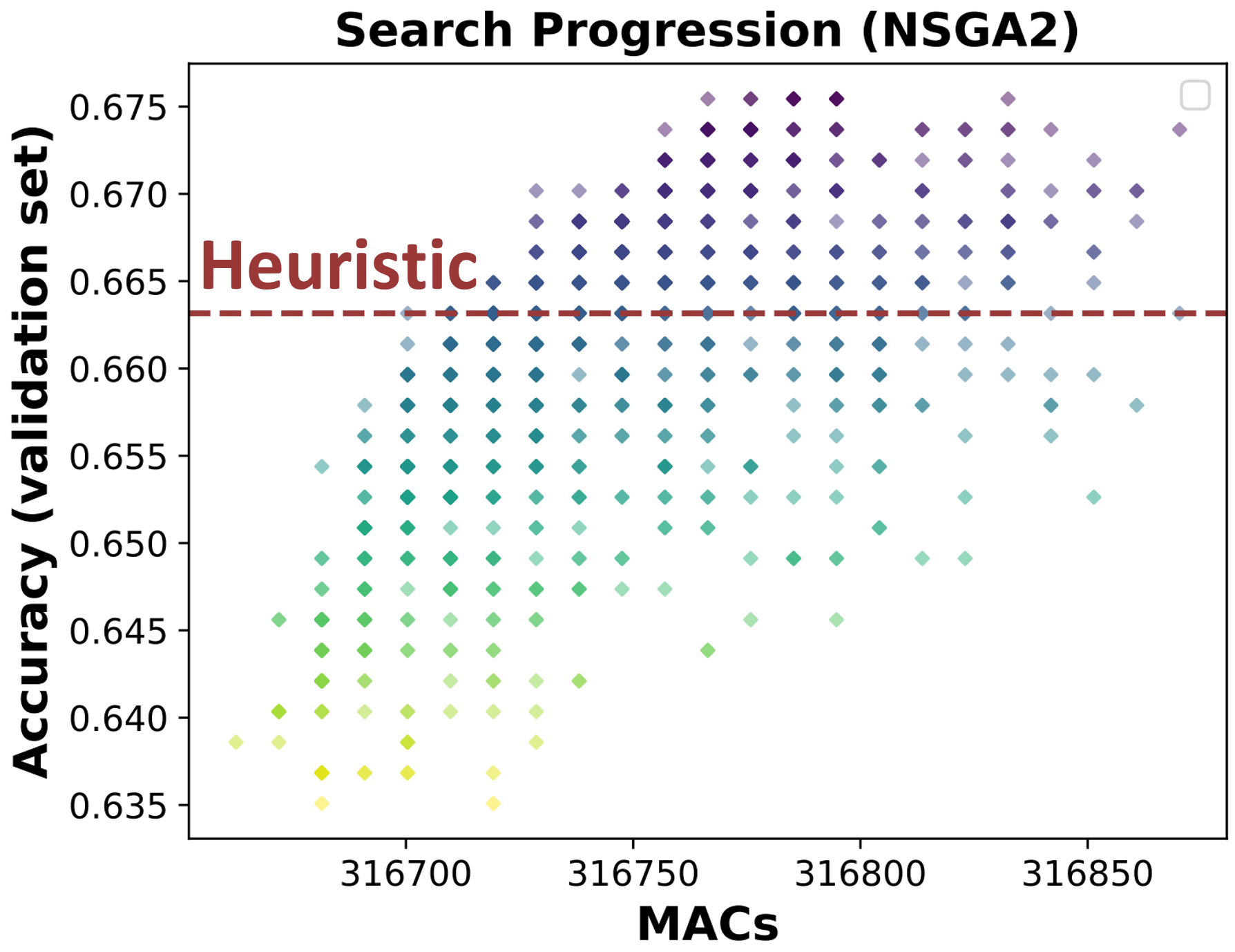}
\caption{Search progression to discover Pareto-optimal low-rank adapter configurations. The horizontal line represents the zero-shot accuracy of the midpoint heuristic sub-adapter.
}
\label{fig:nsgaii}
\end{figure}

The larger research community can further improve the solutions discussed here. For instance, the additional stage to discover a high-performing adapter configuration from the search space of possible configurations presents several opportunities for improvement. As illustrated in Figure \ref{fig:nsgaii}, evolutionary algorithms, e.g., the Non-Dominated Sorting Genetic Algorithm II
(NSGA-II) \cite{nsgaII}, might be used to discover Pareto-optimal elastic low-rank adapter configurations. In this example, a multi-objective search is performed on multiply-accumulate (MAC) operations and validation accuracy. This search can be expensive, presenting opportunities for more efficient alternatives.

\vspace{-0.5cm}
\section{Conclusion}

This retrospective paper discusses recent work on low-rank representations and the synergy with neural architecture search (NAS). The results from the papers that propose solutions aligned with this synergy confirm the benefits in both directions: (i) Low-rank adapters are enhanced by NAS techniques, i.e., elastic low-rank adapters achieve better results than their vanilla low-rank adapter counterparts, and (ii) NAS becomes more efficient by incorporating the utilization of low-rank representations. This synergy also motivates future work to better understand the interaction between these two domains and propose more sophisticated solutions that expand on existing work.

\bibliography{main}

\end{document}